# Automatic post-picking improves particle image detection from Cryo-EM micrographs

Ramin Norousi[1*], Stephan Wickles[2*], Thomas Becker[2], Roland Beckmann[2], Volker J. Schmid[1], Achim Tresch[2]

[1]Department of Statistics, Ludwig-Maximilians-University München, Germany

[2]Center for Integrated Protein Sciences and Munich Center for Advanced Photonics at the Gene Center, Department of Biochemistry, Ludwig-Maximilians-University München, Germany

* both authors contributed equally

Send correspondence to: tresch@lmb.uni-muenchen.de, ramin.norousi@campus.lmu.de, wickles@lmb.uni-muenchen.de

## ABSTRACT

Cryo-electron microscopy (cryo-EM) studies using single particle reconstruction is extensively used to reveal structural information of macromolecular complexes. Aiming at the highest achievable resolution, state of the art electron microscopes acquire thousands of high-quality images. Having collected these data, each single particle must be detected and windowed out. Several fully- or semi-automated approaches have been developed for the selection of particle images from digitized micrographs. However they still require laborious manual post processing, which will become the major bottleneck for next generation of electron microscopes. Instead of focusing on improvements in automated particle selection from micrographs, we propose a post-picking step for classifying small windowed images, which are output by common picking software. A supervised strategy for the classification of windowed micrograph images into particles and non-particles reduces the manual workload by orders of magnitude. The method builds on new powerful image features, and the proper training of an ensemble classifier. A few hundred training samples are enough to achieve a human-like classification performance.

# 1 INTRODUCTION

The 3D cryo-electron microscopy (3DEM) reconstruction is a widely used and powerful technique for achieving structural information from biological macromolecule assemblies. In this approach, thousands of 2D grayscale projections of a macromolecule (referred to as particles) obtained by cryo-electron microscopy (cryo-EM) are used to infer its 3D structure by a 3D alignment (Woolford et al. 2007). The success of the reconstruction crucially depends on the number of 2D images as much as on their quality. Contamination of the dataset with artifacts or noise can lead to severe distortions in the result, including erroneous extra electron densities.

Great efforts were made to accurately pick particle images from low-dose electron micrographs as they were generated by the electron microscope. A variety of excellent methods have been proposed so far. They can be divided into three categories (Zhu 2004): generative, discriminative, and unsupervised approaches. Generative approaches measure the similarity of a candidate image region to a reference molecule. Most of these template-matching techniques use cross-correlation as similarity score (Chen und Grigorieff 2007, Hall 2004, Huang 2004, Roseman 2003). However, these algorithms initially require a representative set of reference 2D projections. Discriminative methods are less demanding, they merely need a training dataset, which contains positive samples (regions containing a particle) and negative samples (regions without a particle). The training set is used to learn a binary classifier. This can be done either in a fully supervised setting (Hall 2004, Mallick S. 2003, Volkmann 2004), or in a semi-supervised, iterated fashion (Sorzano et al. 2009) where the user can correct the algorithm during the training phase. The category of unsupervised approaches entails algorithms that work without any reference. Particles are automatically detected from a micrograph based on statistical measures and features which are extracted from the candidate region (Adiga et al. 2005, Voss et al. 2009, Woolford et al. 2007, Zhu 2004, Ogura und Sato 2005). All these techniques are focused on optimization of particle picking within micrographs. They optimize this step in the sense of detecting of large number of particles from micrographs with as little as possible false positives. The output of these methods is a set of windowed images from micrographs whose quality depends on the signal-to-noise ratio of the micrograph, the picking method and the type of the specimen. Although particle picking methods are invaluable in the automation of the cryo-EM reconstruction process, the fraction of false positives in the output may vary from 10% up to more than 25% (Zhu 2004). In order to improve the 3D reconstruction performance, subsequent manual curation is still inevitable and actually constitutes the major bottleneck for high resolution reconstruction of unsymmetrical molecules. The necessity of further image analysis automation is therefore evident.

Instead of picking particles directly from micrographs, we propose to subject the output of the above methods, which is a mixture of windowed particle images and some windowed non-particle images (such as contaminants and noise), to another round of classification. We realized that the task of identifying spatially constrained objects on a large micrograph image (commonly termed particle picking) is distinct from the task of individually discriminating particles from non-particles in a collection of small images of standardized size ("post-picking"); thus both tasks should be addressed in different steps. To that end, we established Mappos (MAchine learning Algorithm for Particle POSt picking), a supervised, discriminative particle identification method based on suitable features calculated from the candidate windowed images. During the learning phase, a binary classifier model is generated based on the predefined features calculated on the training data. Then, in the detection phase the generated classifier is applied on the rest of the previously unclassified images, which are labeled as either a particle or a non-particle. Prediction accuracy was further improved using an ensemble classifier (Duda et al. 2000). Mappos is a simple yet versatile tool that has been implemented in MATLAB (version 6.0 or greater). It requires the image processing toolbox and the statistics toolbox as add-on packages. The source code for Mappos can be downloaded from the website http://www.tresch.genzentrum.lmu.de/mappos.html and is free for academic use. In terms of resolution and reconstruction accuracy, we show that Mappos performs as good as handpicking, while significantly improving the processing time. The classification performance of our method was evaluated on an artificial dataset as well as on a real FEI Titan Krios electron microscope dataset of the 70S *E. coli* ribosome recorded on an FEI Titan Krios. In this controlled setting, the false positive rate could be reduced by Mappos from 10% after particle picking to 6%. Further Mappos were successfully used to reconstruct a cryo-EM analysis of ribosome recycling complexes (Becker 2012).

## 2  METHODS

Following the standard workflow for a classification task (Bishop 2006), Mappos can be divided into a learning phase, followed by a prediction (particle detection) phase as depicted in Figure 1. The method relies on the availability of a relatively small set of training sample images which have been labeled manually as particles (+) or non-particles (-). This training set should contain a few hundred sample classified images which contain an approximately balanced number of samples of both image types. From each training sample, a vector of numerical features is extracted. A feature is a one-dimensional statistic that is calculated from a sample object. Together with the labels, the feature vectors serve as input to the learning algorithm. We evaluated the performance of several algorithms and decided to use an ensemble of several classification models. The result of the learning phase is a binary classifier C which during the prediction phase assigns a binary label (+/-) to each image from a set of new, unclassified images.

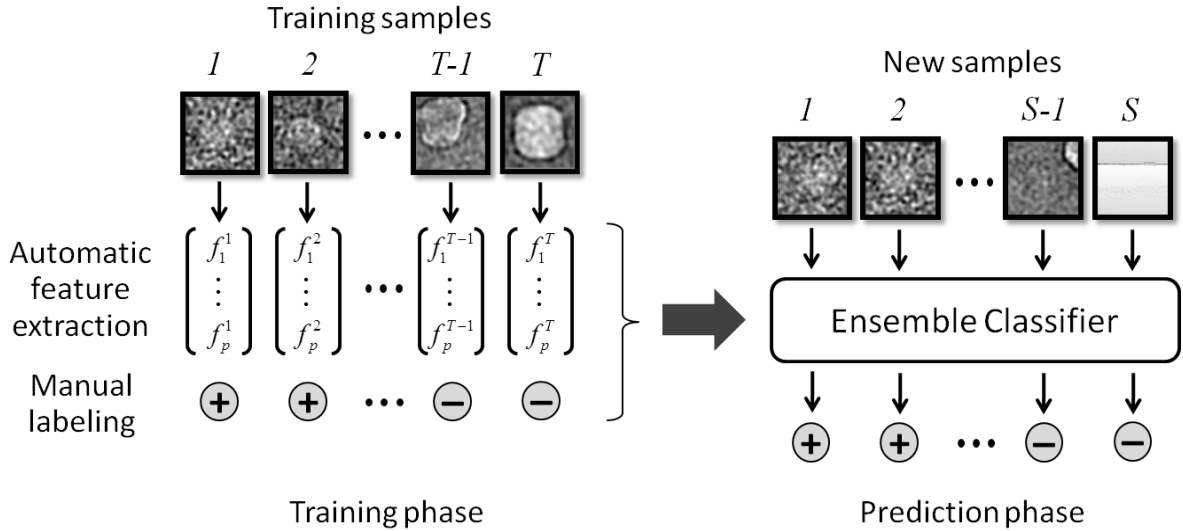

**Figure 1**. Workflow of Mappos. During the training phase, *T* sample images are manually classified as a particle (+) or a non-particle (-). Afterwards, *p* Appropriate features ($f_1^j,...,f_p^j$) are automatically extracted from each sample image *j*. The feature matrix ($f_k^j$), together with the known sample labels are used to train an ensemble classifier. In the prediction phase, the classifier is applied to efficiently label the *S* previously unseen samples.

## 2.1 Discriminatory Features

The success of our method crucially depends on the definition of meaningful features which, as an ensemble, have a good discriminatory power. The screening for suitable features was done on a yet unpublished dataset of cryo-EM images of the *E. coli* ribosome taken by the FEI Titan Krios electron microscope. This dataset contains 1,638 manually classified images, respectively 50% of them representing particles / non-particles. The discriminatory power of a single (continuous) feature is usually assessed by an ROC-curve (Bradley 1997, Fawcett 2004), see also Section 2.3. The area under the ROC curves (AUC) provides a good measure to summarize the performance of the classifier (Langlois und Frank 2011) and ranges between 0.5 (equivalent to random guessing) to 1 (perfect classification). Based on this criterion, we identified the following promising features (AUC values are given in brackets):

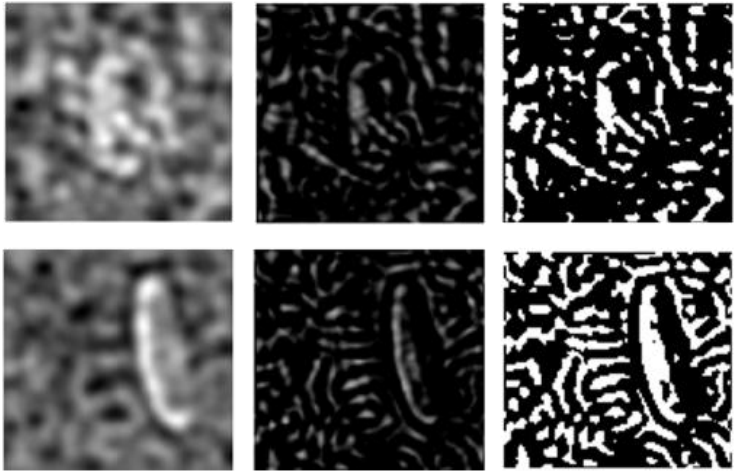

**Figure 2.** Phase symmetry transformation and binarization of a particle image (upper row) and a non-particle image (lower row). Shown are the original particle images (left panel), the output after phase symmetry transformation (middle panel), and the result after its binarization (right panel). The non-particle image contains more regions with a high degree of symmetry. Therefore the image after binarization contains more white pixels as the particle image.

**Radially weighted average Intensity** (0.83). The radially weighted average intensity is calculated as a weighted sum of the pixel intensities, the weights being inversely proportional to their Euclidean distance from the center of the image. This statistic measures the centrality of the bright pixels, which for particle images should exceed that of non-particle images.

**Phase Symmetry / Blob Detection** (0.94). Blob detection (Kovesi 2000) is based on the notion of phase symmetry, a contrast- and rotation invariant measure of local symmetry at each point of an image. Phase symmetry recurs on a 2D Wavelet transformation that extracts local frequency information (Morlet 1982). We apply the phase symmetry transformation with standard parameter settings as in (Kovesi 2000). The transformed image is binarized using Otsu's thresholding (Otsu 1978). Afterwards, locally symmetric areas ("blobs") mainly occurring in non-particle images can be counted. We report the relative frequency of 1's in the binarized picture as a feature (Fig. 2).

**Dark Dot Dispersion** (0.86). We noticed that in particle images, the "dark dots" are distributed more evenly across the image than for non-particle images. After convolution of the image with a 2-dimensional symmetric Gaussian kernel, dark dots are defined as connected regions of intensity less than the 5% quantile of the overall intensity values. The center of a dark dot is calculated as the mean of its pixel coordinates. The dark dot dispersion of an image is defined as the variance (the mean squared Euclidean distance) of its centers. Further helpful features were the (0%,10%,50%,90%,100%)-quantiles of the pixel intensity distribution of an image, the number of foreground pixels after binarization, and the number of edges counted after Canny edge detection (Canny 1986).

## 2.2 Construction of a Classifier

Instead of learning just one classifier, we use an ensemble method, in which the final classification is the majority vote of a number of individual classifiers (Hastie et al. 2009). We implemented an idea described in (Wichard 2009), which combines bagging (Breiman 1996) and cross-validation. Assuming a given set of classified samples is, we hold aside 10% of the classified samples which serve as a validation set which for assessing the performance of the final classifier ensemble. The remaining 90% will be used for generating the classifier ensemble, which takes $K=21$ rounds. In each round the remaining 90% of the data is divided randomly into a training set and a test set with a ratio of 4:1. In every round a collection of classifier models is learned on the training set, and best model, i.e. the model with the lowest classification error on the test set, is added to the classifier ensemble. Finally we obtain an ensemble of 21 classifiers, and the final decision is based on majority vote. We trained and tested different classifiers with a variety of parameter choices. Among the methods we considered were Linear Discriminant Analysis, decision trees, logistic regression, multilayer perceptions, support vector machines, and n-nearest-neighbors (Hastie et al. 2009). We found that in our situation decision trees are superior to other models; hence we finally decided to use decision trees as the basic learning method. Using the held-out 10% of the training data, we report the sensitivity and the specificity associated with the final classifier ensemble.

Our aim was to devise a method containing as few external parameters as possible, because parameter tuning involves the risk of overfitting. Mappos is not parameter-free, since the learning methods internally called by Mappos contain tuning parameters; they are all run with their standard parameter settings as given by MATLAB. The fact that there is no single explicit parameter that needs to be tuned by the user greatly contributes to the robustness of Mappos and its user-friendliness.

## 2.3 Construction of a Training Set and Specificity Tuning

We suggest running Mappos with a hand-picked training set of 500 particle images and 500 non-particle images. If there exist artifacts of different types (e.g., those mentioned in Fig. 3), it is advisable to choose the non-particles evenly from each type. It is good practice to cross-check the final output of Mappos by eye to ensure a sufficiently high specificity of the selection procedure. In case it needs to be increased, the initial training set should be extended by another set of hand-picked non-particle images, 500 say. This process can be iterated; however this was never necessary in our applications.

## 2.4 Validation

The Performance of Mappos was assessed on simulated and real data of the *E. coli* ribosome. For the simulated data, the true labels of the images were known. Standard performance measures were calculated for 2x2 contingency tables of true vs. predicted labels (for their definition, see Table 1). For the real data, the manual classification was taken as a gold standard. We additionally assessed the quality of the electron density map after reconstruction.

## 2.5 Artificial Data

We generated 21,922 windowed images with a particle/non particle ratio of 9/1, which is comparable to that in real cryo-EM data sets. The images for particles and non-particles were generated by projecting 3D volumes evenly distributed into 2D. Making a meaningful statement about the classification performance of our algorithm requires that our model images resembles real cryo-EM pictures in fundamental properties like signal-to-noise ratio (SNR) and image contrast modulated by the contrast transfer function (CTF). This was achieved by the image manipulation procedure described by Frank et al. (Frank 2006). First, the structural noise in real data sets is simulated by adding random noise with zero-mean Gaussian distribution to a SNR of 1.4. Second, the image formation of a bright field microscope working under 300kV and a defocus of 2.0μm was simulated by modulation of the pictures with a contrast transfer function (CTF). The final step was to add random noise (shot and digitization) of zero-mean Gaussian distribution to a SNR of 0.05. By analogy to image processing of real cryo-EM images, the artificial pictures were also low-pass filtered to reduce the noise. The image manipulation workflow is depicted in Fig. 3a. In order to verify that Mappos can cope with all types of contaminations, non-particle images were generated from four 3D templates that served as a projection volume: plate, cylinder, sphere, and void (Fig. 3b). These templates were chosen such that they covered the spectrum of contaminations typically encountered in cryo-EM images (Fig. 3c)

## 2.6 Cryo-EM Data

We compared three methods (i) manual post-picking, (ii) no post processing and (iii) post-picking with Mappos on a real cryo-EM data set of empty 70S ribosomes from *E.coli*. Micrographs were automatically collected on an FEI Titan Krios electron microscope under low dose conditions. Besides their classification performance, we assessed the effect of post-picking on the reconstruction quality of the electron density map. We defined an input data set consisting of 85,726 windowed projection images which were detected by the template matching algorithm of Signature. For automated classification, a training dataset of 2,000 particles (50% particles resp. non-particles) was provided. All data sets were processed using SPIDER and refined for 3 rounds to a final resolution of about a Fourier-Shell Correlation (FSC$_{0.5}$) of 11 Å.

Figure 3. (a) Generation of an artificial cryo-EM image based on a crystal structure of the 70S ribosome (2QAL, 2QAM Cate (2007)). The first image shows a 2D projection of the ribosomal electron density which is as depicted in the next image further modified by adding noise to account for structural heterogeneity. The third picture is additionally CTF-distorted and illustrates the image of a bright field electron microscope. Low dose conditions are simulated by adding noise to a SNR of 0.05. The last picture is band-pass filtered to improve the contrast and served as input for our classification.
(b,c) Comparison of real cryo-EM images (b) and the artificial projections (c). The last column depicts the 3D volumes which are used to generate the artificial data set. Projections of the *E.coli* ribosome in different views are illustrated in the first row. The next four rows show different groups of contaminations commonly found in cryo-EM data sets.

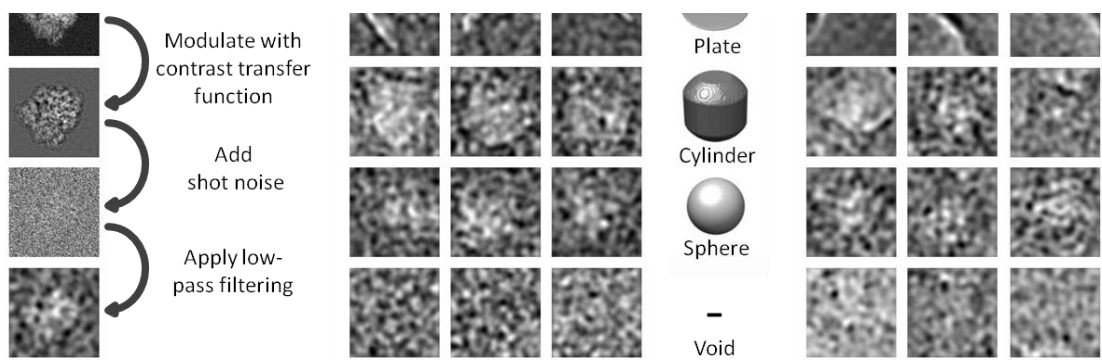

## 3   RESULTS

For the purpose of electron density reconstruction, the effect of false positives is severe. A percentage of 20% of false positives can be detrimental. Our algorithm was therefore designed to yield a particle selection containing as few as possible false positives, i.e., our objective was a high positive predictive value for the cryo-EM data, and a high specificity for the simulated data (the terms are defined in Table 1). Since windowed images are typically available in abundance, we deliberately sacrifice a bit of sensitivity and trade it against a high specificity/positive predictive value of the method.

### 3.4  Artificial data

In order to evaluate the performance of Mappos in a controlled simulation environment, we generated artificial data sets as follows: The training set was composed of 1,000 positive and 1,000 negative samples. Another 19,874 positive and 2048 negative samples served as a test set. The positive samples were a selection from random particle projections of the *E.coli* ribosome. We investigated five scenarios, depending on whether the negative samples were drawn exclusively from Plate/ Cylinder/Sphere/Void projections (see Fig.3 and Methods 2.5), or from a uniform mixture of all of them. The result of one simulation run is a vector (TP,FP,TN,FN) of true positives, false positives, true negatives, and false negatives. This outcome was converted into specificity and sensitivity values, as defined and reported in Table 1. The proportion of negative and positive samples can be chosen arbitrarily in our simulations. Sensitivity and specificity of Mappos are therefore the most informative figures, because they do not depend on the ratio of particle to non-particle samples. We are able to calculate the mean and the variance of both performance measures by running 100 independent simulations in each scenario (Table 1). It turns out that the specificity is above 70% in all scenarios, while the sensitivity always exceeds 80%. Roughly speaking, we have shown that post picking with Mappos reduces the number of non-particles by a factor of 5, while reducing the number of true particles only by a factor of 1.25. This enrichment of particles in set of post-picked images can lead to substantial improvements in the 3D reconstruction.

| Conta-mination type | TP ±relative std.dev. | FP ±relative std.dev. | TN ±relative std.dev. | FN ±relative std.dev. | **Sensitivity ±std.dev in %points** | **Specificity ±std.dev in %points** |
|---|---|---|---|---|---|---|
| Plate | 814 ±2% | 300 ±6% | 701 ±2% | 186 ±10% | **81% ±2%** | **70% ±2%** |
| Cylinder | 865 ±1% | 144 ±10% | 856 ±2% | 201 ±9% | **87% ±1%** | **86% ±1%** |
| Sphere | 798 ±2% | 241 ±8% | 759 ±3% | 202 ±9% | **80% ±2%** | **76% ±2%** |
| Void | 806 ±3% | 296 ±7% | 704 ±3% | 194 ±12% | **81% ±2%** | **70% ±2%** |
| All | 794 ±2% | 257 ±7% | 743 ±2% | 206 ±9% | **79% ±2%** | **74% ±2%** |

**Table 1.** Performance of Mappos in 5 test scenarios. A set of 700 particles and 700 non-particles was used for training in each case. The test set consisted of 1000 particles and 1000 non-particles. For self-containedness, we provide a definition of the performance scores proposed by (Langlois und Frank 2011) for the comparison of particle picking methods. Classification results on the test set were compared to the known labels, splitting the samples into correctly classified particles (true positives, TP) resp. non-particles (true negatives, TN), and samples incorrectly classified as particles (false positives, FP) resp. non-particles (false negatives, FN). Quantities that are derived from these four numbers are the sensitivity, Sens.=TP/(TP+FN), specificity, Spec.=TN/(TN+FP), positive predictive value, PPV=TP/(TP+FP), negative predictive value, NPV=TN/(TN+FN), and accuracy, (TP+TN)/(TP+TN+FP+FN). Sensitivity and PPV are highlighted in bold print, because in our setting, it is particularly important to eliminate non-particles from the result set (corresponding to a high PPV), while still recovering a sufficient number of true particles (corresponding to a satisfactory sensitivity).

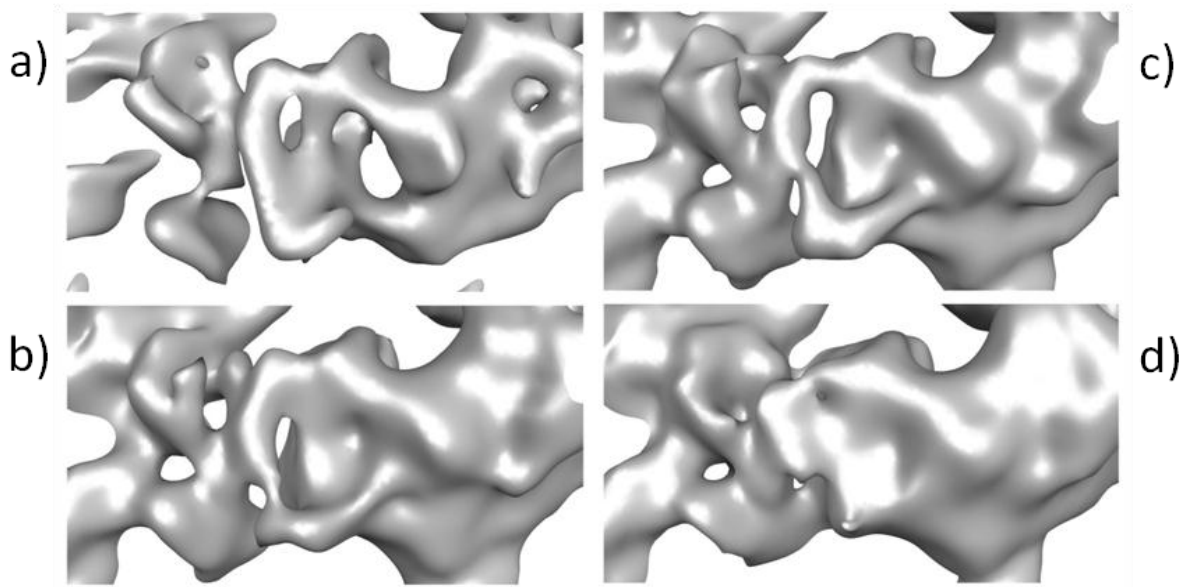

**Figure 4.** Electron density maps calculated from one differently classified real cryo-EM data set of the *E.coli* 70S ribosome. Shown is a close-up of the ribosomal tunnel exit with protein L29. Pictures from top to bottom: a) electron density map based on the crystal structure (2QAL, 2QAM, Cate (2007)) at a resolution of 10Å, b) and c), reconstruction originating from the manually respectively the automatically classified data set, d) reconstruction based on the unclassified data set.

### 3.5 Cryo-EM data

Particles were picked using the template matching software Signature (Chen und Grigorieff 2007) and classified by Mappos. The same particles were manually inspected and classified by a human expert, thus creating a gold standard. Expectedly, both sensitivity and PPV increase with the size of the training set. Note that the moderate increase in PPV (from 86% in the input data set to 93% after post-picking) corresponds to a substantial reduction of the fraction of false positives by a factor of 2.5 (Figure 5). As an additional criterion, we consider the quality of the electron density map in terms of structural features that appear clearly resolved. A map of the unclassified data set serves as a negative control. In Figure 4, the ribosomal tunnel exit with protein L29 is shown as close-up to determine the quality of the density map. According to Fourier-Shell correlation, the resolution of the reconstruction was comparable in all cases, but there were clear differences in the quality of the density maps, in terms of structural features that are clearly separated in the reconstruction. Having the ribosome as molecule, the first features to be resolved are RNA helices followed by protein α-helices and β-sheets. In some cases it is also possible to resolve bulky side chains and linker regions. The density map of the unclassified data set shows no clear separation between ribosomal RNA (rRNA) and ribosomal proteins (rp). It is also not possible to extract secondary structure information. Only the overall shape of the tunnel exit can be extracted from this map. The reconstruction of the manually inspected data set contains information about the localisation and secondary structure of the proteins. The a-helices of rpL29 and rpL23 are almost completely resolved. There is no clear separation between parts of the ribosome where proteins interact with large RNA structures. The reconstruction of the automatically classified data set also shows structural features of ribosomal proteins. The electron density of one a-helix of rpL29 as well as the linker region between the two helices is already resolved.

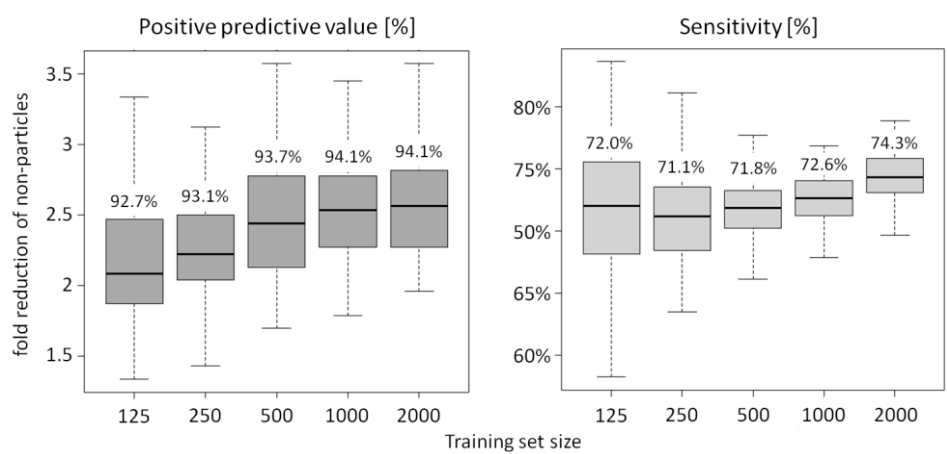

**Figure 5**. The performance of Mappos on the ribosome data set in terms of positive predictive value (left panel) and sensitivity (right panel) as a function of the training set size (x-axis). The boxes summarize the results of 100 replicate *in silico* experiments, their median is given on top of the box. In the left panel, the reduction in the number of non-particles in the output of Mappos (false positives) relative to the number of non-particles in the input set is given as a factor (y-axis).

## 4 DISCUSSION

We introduced Mappos, an ultra-fast particle picking method that reduces the amount of required manual preprocessing by orders of magnitude, while achieving a comparable specificity and sensitivity as for manual picking. We demonstrated at the example of the *E. coli* ribosome that the quality of the electron density map after reconstruction from the set of automatically picked particles is identical to that of a hand-picked particle set. Existing particle picking methods like (Sorzano et al. 2009) aimed at the simultaneous identification and classification of "true" particles on a micrograph. One of our very simple yet key findings is that these two tasks should be addressed separately. The reason is that in the post-picking step, expert knowledge about the characteristics of (non-)particles can be incorporated into the automated classification with very low effort. In particular, we can present positive *and* negative samples of windowed images to our learning procedure, whereas conventional picking approaches use only positive samples as a reference, focusing on the detection of particles. The availability of negative samples contributes greatly to the specificity of Mappos. Consequently, the focus of the particle identification step can be put on sensitivity (filter criteria can be less strict), because a sufficient specificity is ensured by the subsequent

Mappos step. Note that it would be difficult even for human experts to achieve an inter-observer agreement of more than 90%. Therefore, we feel that Mappos, achieving a specificity of 94%, performs almost optimal on our example.

The definition of suitable image features turned out to be crucial to the method's success. Once discriminative features are found, the training of a classifier (ensemble) is relatively easy, and many black box learning algorithms perform comparably well. Obviously, there is always room for improvements in the definition of additional informative features. One might think of deriving features that specifically detect a certain class of non-particles.

Furthermore, it would be desirable to automatically tune the specificity of Mappos to avoid a potential second round of training and classification. This is in theory feasible, but it would not relieve the user from the duty of checking the final result by eye, and an automatic tuning would bear the danger of decreasing the sensitivity of Mappos excessively. We therefore rather recommend enriching the training set with negative examples in this case.

So far, the power of our method has been proven only for one class of particles, the *E. coli* ribosome. It remains to show its applicability to a wider class of particles. Recently, Mappos has successfully been used to reconstruct a cryo-EM analysis of ribosome recycling complexes (Becker 2012). With regard to the huge amount of data being generated by the current generation of electron microscopes (The speed of data acquisition on a Titan Krios EM is 2.000 micrographs per day, which amounts to ~200.000 particles per day), automated particle picking tools will inevitably become an integral part of every cryo-EM pipeline. Hand-picking of the E.coli data set requires many working days. Mappos is extremely fast, it can handle this deluge of data at the same pace at which it is generated. E.g., it took 1 hour for one person to generate the training set, and approx. 2 hours to run Mappos on a standard desktop computer. As we have demonstrated, the quality of the final 3D reconstruction equals that of the hand-picked data set, although the quality of the raw data was moderate (see Fig. 4x) and required substantial filtering. The utility of Mappos is presumably maximal for unsymmetrical, large molecules, for which a large number of particles needs to be picked for high resolution cryo Electron Microscopy.

**Authors' contributions**

Achim Tresch and Roland Beckmann initiated the research. Ramin Norousi, Volker Schmid, and Achim Tresch developed the algorithm. Stephan Wickles, Ramin Norousi, Thomas Becker and Achim Tresch analyzed the data. Achim Tresch, Ramin Norousi and Stephan Wickles wrote the manuscript.

**Acknowledgements**

This work was supported by the Deutsche Forschungsgemeinschaft (SFB646). Achim Tresch was supported by a Gastprofessur grant from the Deutsche Forschungsgemeinschaft to Patrick Cramer.